RESEARCH ARTICLE                                         OPEN ACCESS

# Customer Churn Prediction Model using Explainable Machine learning

**Jitendra Maan [1], Harsh Maan [2]**

[1] Head - AI and Cognitive Experience, Tata Consultancy Services Ltd. - India
[2] DWH/BI Developer, Amdocs - India

**ABSTRACT**
It becomes a significant challenge to predict customer behaviour and retain an existing customer with the rapid growth of digitization which opens up more opportunities for customers to choose from subscription-based products and services model. Since the cost of acquiring a new customer is five-times higher than retaining an existing customer, henceforth, there is a need to address the customer churn problem which is a major threat across the Industries. Considering direct impact on revenues, companies identify the factors that increases the customer churn rate. Here, key objective of the paper is to develop a unique Customer churn prediction model which can help to predict potential customers who are most likely to churn and such early warnings can help to take corrective measures to retain them. Here, we evaluated and analysed the performance of various tree-based machine learning approaches and algorithms and identified the Extreme Gradient Boosting "XGBOOST" Classifier as the most optimal solution to Customer churn problem. To deal with such real-world problems, Paper emphasize the Model interpretability which is an important metric to help customers to understand how Churn Prediction Model is making predictions. In order to improve Model explainability and transparency, paper proposed a novel approach to calculate Shapley values for possible combination of features to explain which features are the most important/relevant features for a model to become highly interpretable, transparent and explainable to potential customers.

## I. INTRODUCTION

Customer Churn has become an industry-wise problem due to unprecedented competition in post pandemic world. In today's world of rapidly changing business environment, it has become customary to develop innovative methods and approaches to predict customer who would likely to churn. According to Umayaparvathi et al. [9], churn prediction is important to identify the possible churners in advance to do corrective actions.

Therefore, it has become important to identify important features in the dataset who may likely contribute to higher Customer churn to take appropriate action well in time.

The paper explained the detailed methodology used to evaluate the accuracy and performance of various machine learning ensemble models (like, Logistic Regression, Random Forest, Decision Tree and Extreme Gradient Boosting "XGBOOST") and then select one of the most optimal model to address the issue.

In today's changing business environment, it is essential to trust the outcome of such Customer Churn prediction Models whereas explainability and transparency is of a major concerns identified by Customers across business domains. Such ethical concerns are rapidly growing with little or no visibility on explanation of how machine learning models predicted the outcome. Sensing such big void due to lack of innovative solution to address such global pervasive problem, henceforth, there is a dire need to design, develop and deploy machine learning models which are ethical in their purpose, design and usage covering key aspects of transparency, explainability and interpretability.

Customer Churn Prediction Model is trained with sufficient dataset to generalize and accurately predict customer churn rate for different customers across various industries, segments and business domains.

The overall objective behind such problem statement is to develop Customer Churn Prediction Model which not only predict customer behaviour but also provide insights on various factors that impact the churn rate. The proposed model is trained with sufficient dataset to generalize and accurately predict customer churn rate for different customers across various industries, segments and business domains.

## II. KEY CHALLENGES

In today's competitive business environment, there are plethora of challenges for companies to predict the reasons for customers to leave. A few reasons as highlighted below –

- Comparable services available from competition at much more competitive rates
- Lack of a unified machine learning prediction model to provide triggers for frequent customer churn

- Lack of BI tools and platform to provide visibility into Customer pain areas
- Digitization opened up more choices for customers with subscription based products and services
- Accidental churn, at times, may occur due to sudden change in customer financial situation
- Lack of Customer retention strategies and incentive plans to minimize churn

Looking into above factors, there is a significant need to develop a unique classifier which can accurately predict future churns based on historical data and customer behaviour patterns.

## III. BACKGROUND STUDY AND RELATED WORKS

A. *Literature Survey*

There has been a few papers already published which depict the work done in this field earlier. Although, some new approaches for Customer churn prediction have come out in these papers. It's a quite an emerging field with different machine learning approaches proposed to address the customer churn issue which is quite pervasive across various industry verticals. Hence, our work aims at proposing a novel optimized approach to Customer churn prediction by augmenting the Modeling approach supported by model interpretability and explainability.

Burez and Van den Poel [2], in their paper clearly articulated two different methods on how to manage customer churn in both reactive and proactive way. In a proactive approach, organizations use technology-based approaches to find customers who are likely to churn and develop incentive based strategies to retain them.

Ning Lu [3] has published his research work and proposed the boosting algorithms for customer churn prediction model where different weight are assigned by the boosting algorithm based on customers who are segregated into two clusters. Here, Logistic regression is used as a base learner. His experimental analysis revealed that boosting algorithm provides much better results as compared to single logistic regression model.

There is another interesting research work published by Benlan He [4] who has recommended Support Vector Machine (SVM) model for customer churn prediction and he also used random sampling technique for imbalanced data of customer data sets.

There is another paper titled "Customer churn prediction using improved balanced random forests" by Y.Xie et al., [5] leveraged an improved balance random forest (IBFR) model which combines both balanced random forests and weighted random forests to address data distribution problem. During the experiments, it was observed that IBRF is better than Decision tree, SVM and artificial neural network (ANN), in terms of accuracy.

Makhtar [6] proposed the churn model using set theory where Rough Set classification algorithm has provided better results than Linear Regression, Decision Tree, and Voted Perception Neural Network.

Van Wezel & Potharst [7] Projected an interesting finding thata ensemble learning models provide better accuracy as compared to individual models, whereas Lalwani et al. [8] also concluded that XGBoost and AdaBoost, the boosting classifiers performed better on training and test data with much better accuracy results.

Rothenbuhler et al. [11], studied the churn prediction using Hidden Markov's model based on a stochastic process.

Amin et al. [12] believes that churn prediction and prevention is important for company's reputation which may also impact on revenues.

Most of the previous research work did not build features from raw data whereas, they have more dependency on ready to use features provided by vendors and most of these papers did not inherently implemented the model explainability techniques to provide transparency and model interpretability.

## IV. DATASET FOR TRAINING AND ANALYSIS

There are various datasets available for Customer transactions maintained by Mobile Operators. Current analysis and implementation is based on dataset taken from telecom industry to analyse Customer churn rate.

A sample dataset has been obtained from public domain for Customer churn prediction. Dataset includes various customer profiles and the features spread across various columns covering account level details and subscription plans with day/night call charges and so on. The diagram below provides a glimpse of various featured columns in the dataset.

Fig. 1. Dataset Fields

## V. METHODOLOGY

In this paper, the authors attempted and explored the innovative approach and solution to address the current challenges in predicting customer churn rate based on their experience in more or less similar context. In machine learning parlance, there are solutions where multiple algorithms may fit to the problem domain and most optimal solution is proposed based on comparison of their performance evaluation metrics. However, the solution is conceptually generalized but its implementation takes a unique approach considering technology advancements.

We prioritize customers by taking into account of their state and account length for each of them and confidence of churn prediction.

A. *Data Pre-processing*

Data is initially prepared by aggregating transactional data at each customer level. Post that, we performed exploratory data analysis to better understand the churn data in terms of services, charges, customer plans and tenure and so on. The following tasks are performed in Data Pre-Processing phase of machine learning models:
   a. **Missing value analysis**: During analysis, it is observed that there are no missing value present in the dataset.
   b. **Data Type conversion**: For efficient model development, data type is appropriately converted from Object data type (Boolean) to numerical data types, say for example, Churn, Vmail Plan and Intl Plan are converted into the numerical data type to analyse and process by different classifiers.
   c. **Data redundancy check**: We have also checked for duplication of customer phone no. to ensure the uniqueness of the dataset.

B. *Feature Extraction & Selection*

Feature engineering techniques are used to extract important features from the sample dataset. There are several featured columns in the dataset which are not relevant for Model building and prediction of Customer churn, have been dropped, say for example, Phone, State, Day Charges, Evening Charges, Night Charges, Intl Charges, Vmail Plan, Vmail Message and so on.
There are several methods for feature selection, say for example, recursive feature elimination [10]. Due to time constraint, we are not using such methods, which may be leveraged in future work.

C. *Train-Test Data Split*

For effective model training and testing, the dataset is split into training data (70%) and test dataset (30%) so that data in training set and testing set are unique and the proposed model would be trained over training dataset and it is further tested with unforeseen data from test dataset.

D. *Metrics for Model Evaluation*

In order to access the performance of various classifiers, the following evaluation metrics are leveraged:
- **Feature weights**: It shows an important features used by the model to generate the predictions
- **Confusion matrix:** It depicts a 2*2 metrics of true and false predictions compared to the actual values

- **Accuracy score:** It calculates the overall accuracy of the model for training and test datasets
- **Precision, Recall and Support Metrics:** It depicts the trade-off between precision and recall for different threshold where precision can be seen as a measure of quality, and recall as a measure of quantity.
- **F1 Score:** It is normally the harmonic mean of precision and recall and thereby measures the percentage of correct predictions that a machine learning model has made and it is more suitable for imbalanced data.
- **ROC Curve:** An evaluation metrics for binary classification problem and it clearly signifies if model is capable of distinguishing between classes.

## VI. SOLUTION APPROACH

The proposed machine learning based modelling approach is unique as we developed a highly performance optimized and highly responsible churn prediction system which enforce trust and improve transparency through interpretability of the Model which clearly explain how it took a decision to arrive if customer will churn or not. The below diagram depicts various dimensions of solution uniqueness and novelty.

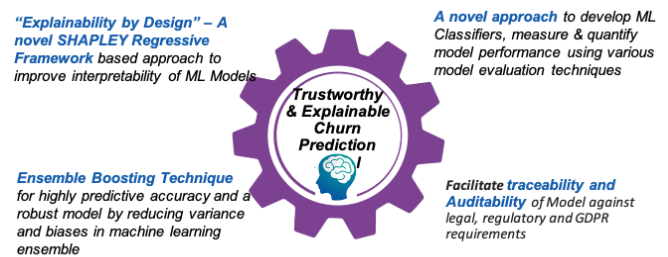

Fig. 2 Key Solution elements

## VII. EXPERIMENTAL RESULTS

A. *Performance Analysis and Key Metrics*

Performance metrics for our proposed model would evaluate the efficiency of various classifiers which is illustrated in below diagram

I.A.1 *Evaluation Metrics*

The diagram below depicts the F1-score for those features which are more important in prediction of customer churn rate leveraging ensemble classifier XG Boost.

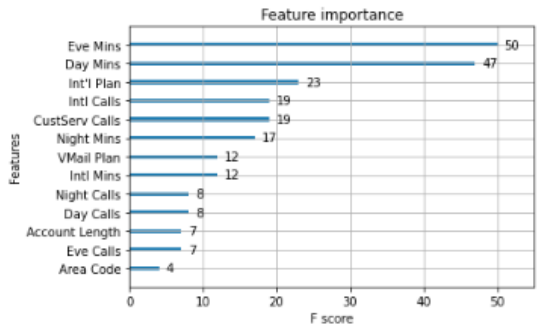

Fig. 3 XG Boost Classifier: Feature Importance and F1 score

The diagram below depicts the Receiver Operating Characteristic (ROC) AUC Curve which triggers the model to optimize the true positive (Correctly predicted churners) and false positive rates (Wrongly predicted churners). This metrics play an important role since we need confidence in our positive class predictions i.e. Churn once we take specific retention actions whereas, accuracy as another evaluation metrics does not assure the same.

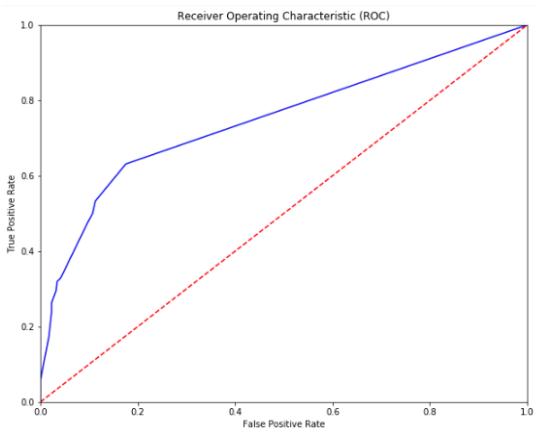

Fig. 4 ROC Curve for XG Boost Classifier

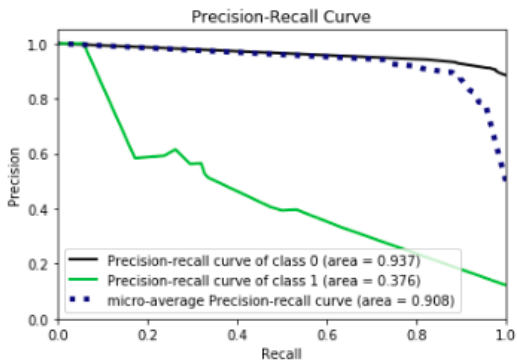

Fig. 5. Precision Recall Curve

The table below provides the **classification Report,** another performance evaluation metrics which is used to show the precision, recall, F1 Score, and support of trained classification model.

Table 1: Classification Report

| Description | Precision | Recall | F1-score | Support |
|---|---|---|---|---|
| Not Churn | 0.9 | 0.97 | 0.94 | 878 |
| Churn | 0.55 | 0.22 | 0.32 | 122 |
| Accuracy | | | 0.88 | 1000 |
| Macro avg. | 0.73 | 0.6 | 0.63 | 1000 |
| Weighted avg. | 0.86 | 0.88 | 0.86 | 1000 |

*I.A.2  Analysis of Churn Prediction Classifiers*

Here, in the proposed study, we analyzed various Classifiers and compared them based on their accuracy and performance to correctly predict Customer churn rate. Once model output is obtained, then proposed study recommends the most optimal Classifier based on various performance metrics/KPIs –

a. Logistic Regression,
b. Random Forest
c. Decision Tree and
d. Extreme Gradient Boosting "XGBOOST"

Machine learning model is developed and trained with these classifiers and then F1 score is calculated as per below table.

Table 2: Accuracy and F1-score of various Classifiers

| S.No. | Classifier Name | Accuracy Score | F1 Score |
|---|---|---|---|
| 1 | Logistic Regression | 0.878 | 0.153 |
| 2 | Decision Tree | 0.922 | 0.614 |
| 3 | Random Forest | 0.954 | 0.796 |
| 4 | XG Boost Classifier | **0.962** | **0.836** |

Based on performance comparison of above classifiers, it was revealed that XGBOOST Classifier provided the highest F1 score and Accuracy score than other 3 models, thereby depicting the best performance among all classifiers. XGBoost ensemble model has the highest AUC of 0.79 with a recall of 0.83 and precision of 0.54.

In order to predict binary churn outcome using XGBoost model, the following scaling parameters are tuned and applied to 16 features as they will impact on model outcome –

a. Objective set: "binary:logistic"
b. colsample_bylevel (per split per level) = 0.7
c. colsample_bytree: 0.8 [range 0.5 ~ 1.0)
d. Learning rate: 0.15
e. Gamma: 1 (default:0) (Depends on loss function)
f. Maximum depth (tree): 4 (In order to avoid overfitting of model, this value has been kept reasonably low which will make the model simple)
g. max_delta_step: 3
h. min_child_weight: 1 (High value lead to model under-fitting)

i. n_estimators: 50
j. Reg_lambda: 10 (L2 Regularization to reduce model Overfitting)
k. Scale_pos_weight: 1.5 (Considering Class imbalance)
l. Subsample: 0.9
m. Silent = False. and,
n. n_jobs: 4

Note: max_depth and min_child_weight parameters in XGBoost model are tuned with appropriate values since they have much higher impact on Model outcome.

The proposed model is recursively trained with different hyper parameters until cross validation errors are reduced significantly.

### I.A.3  Model Explainability

Most of the machine learning algorithms are kind of black box where key stakeholders do not get any clarity on how the machine learning model arrived at certain outcome, henceforth, there is a need to bring more transparency by improving model explainability and interpretability. Henceforth, our study explained the Churn model using Shapley values [13] of the model and the other on the importance of the features [14]

Here, we proposed the SHAP (SHapley Additive exPlanations) Framework which is used to explain/interpret the output of machine learning models.

Our proposed solution is based on XGBoost model, an ensemble tree model, henceforth, we are leveraging TreeExplainer to explain the model outcome.

The below Bar graph represents the mean absolute value of the SHAP values for each important feature.

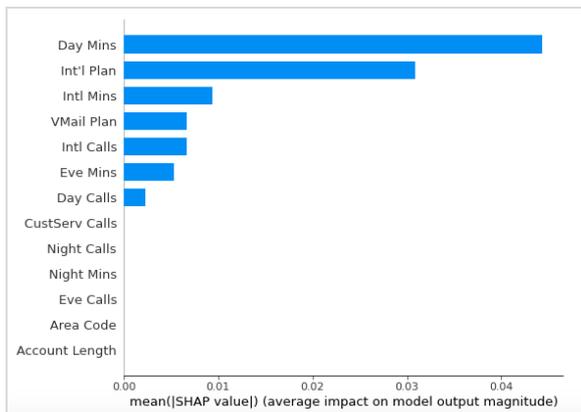

Fig. 6 Force Plot Graph for SHAP Value

The graph below depicts the Shapley (SHAP) values on the x-axis whereas Y-axis represents the important feature values. Values towards the left of centre line represent the observations which move the predicted value in the negative side whereas, the points marked towards the right shifts the model prediction towards the positive side. The below plot is called a force plot and here, features pushing the prediction higher are shown in red color while features pushing it lower shown in blue Color.

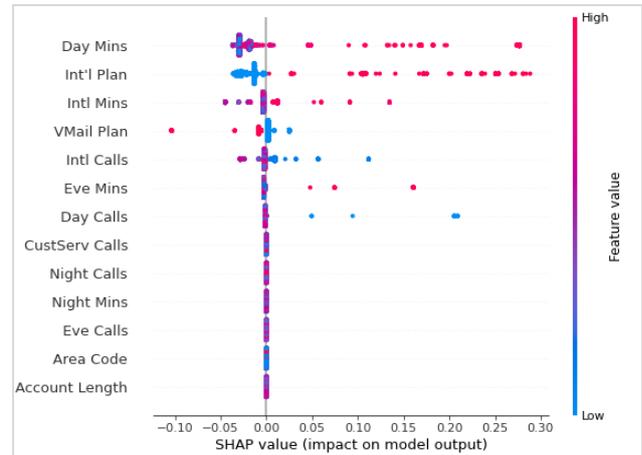

Fig. 7 SHAP Value (Impact on Model Output)

Model Interpretability and Explainability not only improve Customer confidence but also improves customer retention rate. Such standard automation approach with Machine learning models with Test Driven Design approach helps to achieve accurate predictions early on in Customer interaction journey.

## VIII. CONCLUSIONS AND FUTURE WORK

The paper aims to find the most accurate prediction model for Customer churn considering the key reasons that lead to customers churn. The cost of acquiring new customer is way higher than the cost of retaining an existing customer, it is the need of the hour to explore and apply most effective data mining techniques, specifically, machine learning based modeling techniques to predict customer churn rate to retain them. The proposed churn prediction model can be used in other related customer use cases specifically, cross-selling, up-selling as well as customer acquisition.

Experimental results clearly demonstrated that XGBoost tree model achieved the best results across all performance measurement metrics. In continuation, we would predict through hybrid of multiple classifiers for higher prediction accuracy.

Next step is to augment the existing dataset with sensitive attributes (Age, Gender, Sex etc.) and mitigate biases using Adversarial Neural Network to enforce predictions independent of sensitive attributes. Another area for further investigation would be, to explore and analyze other methods/techniques used for feature selection and sampling and their impact on the model outcome.